# Keyphrase Extraction: Enhancing Lists

**Mario JARMASZ and Caroline BARRIÈRE**
Language Technologies Research Centre
Interactive Language Technologies Group, National Research Council of Canada
Gatineau, Québec, Canada, K1A 0R6
{Mario.Jarmasz, Caroline.Barriere}@cnrc-nrc.gc.ca

**Abstract**

This paper proposes some modest improvements to Extractor, a state-of-the-art keyphrase extraction system, by using a terabyte-sized corpus to estimate the informativeness and semantic similarity of keyphrases. We present two techniques to improve the organization and remove outliers of lists of keyphrases. The first is a simple ordering according to their occurrences in the corpus; the second is clustering according to semantic similarity. Evaluation issues are discussed. We present a novel technique of comparing extracted keyphrases to a gold standard which relies on semantic similarity rather than string matching or an evaluation involving human judges.

**Keyphrases:** Keyphrase extraction, clustering, semantic similarity, corpus linguistics, keyphrase evaluation.

## 1 Introduction

A list of keyphrases (words and nominal compounds of great significance in a text) is a good starting point for an alternative representation of documents. Indeed, scientific articles are often accompanied by these to help the reader decide if the article is pertinent, and newspaper headlines can be viewed as a particular form of keyphrases which consist of verb phrases as well as noun phrases. It is arguable that a set of keyphrases alone can effectively characterize a text. This paper explores our intuition that an organized set of keyphrases is much more informative than a list presented in the order of occurrence in the source text or according to an automatic system's confidence.

This research relies on Extractor (Turney, 1999) for the keyphrase extraction process. The resulting lists are usually presented in no apparent order; they are in fact sorted according to the system's confidence. As a refinement, we suggest to present them in progression according to their informativeness. Based on information theory (Shannon, 1948), the information content of a concept $c$ is the negative log likelihood, $-\log p(c)$, where $p(c)$ is the probability of encountering an instance of concept $c$. As the probability increases, informativeness decreases. Therefore a general concept is more frequent than a specific one. We use the Waterloo MultiText System with a corpus of about one terabyte of unlabeled text (Clarke et al. 1995; Clarke and Cormak, 2000; Terra and Clarke, 2003) to approximate the information content of a keyphrase. We estimate informativeness by counting in the corpus the number of documents in which a keyphrase occurs. This is adequate as it gives the same ordering as the negative log likelihood.

Keyphrase ordering according to the progression of informativeness is a valid assumption if a document is about a single topic and that all of the extracted keyphrases are relevant to this topic. In such a case it would be possible to sort all the extracted keyphrases from most general to most specific. This is not always the case in practice. We therefore find it necessary to cluster keyphrases according to their topics and to identify outliers. It has been shown that topics in a document are identified by cohesive regions and that semantic similarity is a good indicator of this cohesiveness (Halliday and Hasan, 1976; Morris and Hirst, 1991). We estimate similarity by using a measure called Pointwise Mutual Information (PMI) (Turney, 2001) which relies on probabilities estimated using again the Waterloo MultiText System. The advantage of using PMI as a similarity measure, beyond the fact that it is a very effective measure (*ibid.*; Terra and Clarke, 2003) is that we can develop a robust system that deals with real-world problems since the measure can return a judgment for virtually any pairs of words and phrases.

The evaluation of the quality of a set of keyphrases is an intricate and subjective task (Barker and Cornacchia, 2000). The standard evaluation technique is to compare the overlap between the set of automatically identified keyphrases and a list of human generated ones (Turney, 2003; Frank *et al.*, 1999; Witten *et al.*, 1999). This is problematic as an author will often specify keyphrases that are not found in his article, but keyphrase extraction systems will only select words found in the document. We propose an evaluation method which calculates the overlap by measuring the semantic similarity rather than string matches. This approach will not allow evaluating whether our re-ordering of keyphrases actually helps a reader to understand the gist of a text, but it will allow an evaluation of the usefullness of reordering and clustering to identify outliers.

We begin by discussing related work on keyphrase extraction, semantic similarity, lexical cohesion and the evaluation of keyphrases. Section three describes the collection of texts used for our experiments, the keyphrase extraction process, and the proposed manner for re-ordering the resulting keyphrases. Some results and an evaluation method are discussed. Section four presents the keyphrase clustering algorithm, as well as some results and an analysis of the usefulness of such a post-processing step. Section five presents our conclusions and future work items.

## 2   Related Work

Since our research is a refinement of an existing keyphrase extraction system, it is important to begin with a study of the state-of-the-art. We follow this by briefly discussing metrics for evaluating semantic similarity, which is the foundation of our clustering approach. We then present the notion of lexical cohesion as an indicator of subtopics in a text. Finally we examine keyphrase evaluation strategies.

### 2.1   Keyphrase Extraction

The task of extracting keyphrases from a text consists in selecting salient words and multi-word units, generally noun compounds no longer than five words, from an input document. This is different from keyphrase assignment, where keywords are selected from a list of *controlled index terms* (Witten *et al.*, 1999; Turney, 2003). A survey of automatic techniques for extracting important phrases from a text is presented in (Turney, 1999). Although there are several methods for automatically extracting keyphrases, we find three systems to be of particular interest. These are Extractor (*ibid.*), Kea (Frank *et al.*, 1999; Witten *et al.*, 1999) and NPSeeker (Barker and Cornacchia, 2000).

Extractor requires a corpus for training a genetic algorithm to fine-tune a set of parameterized heuristics. The system performs well for various types of documents and is generally considered to be state-of-the-art in keyphrase extraction systems (Turney, 1999).

The Kea system uses TF×IDF (term frequency × inverse document frequency) and the distance from the beginning of the document as features for learning which keyphrases to extract. In this manner, the keyphrases which appear frequently in the document, but are rare in the corpus, as well as those that appear at the beginning of the document are most probable to be extracted. Kea uses a Naïve Bayes classifier to assign weights to features. It can be trained very quickly, and therefore adapted to specific contexts easily (Frank *et al.*, 1999).

The authors of Extractor and Kea agree that in experimental evaluations, using independent testing corpora, both keyphrase extraction systems achieve roughly the same level of performance, measured by the average number of matches between author-assigned keyphrases and machine-extracted keyphrases (Turney, 2003; Frank *et al.*, 1999; Witten *et al.*, 1999).

To extract keyphrases, NPSeeker performs the following operations: it skims the document for base noun phrases; it assigns scores to noun phrases based on frequency and length; finally it filters some noise from the set of top scoring keyphrases. Barker and Cornacchia (2000) argue that although NPSeeker and Extractor perform differently, human judges have established that the resulting keyphrases are of similar quality. According to their evaluation, this simple system performs no worse than the state-of-the-art.

Extractor, Kea and NPSeeker have been used as components of a configurable Text Summarization system (Copeck *et al.*, 2002). The authors used machine learning techniques to discover the best parameters of their system. In their case Extractor seems to perform best. Using these results as a reference, and due to the thorough testing of Extractor by its author (Turney, 1999), we decided to use this keyphrase extraction module for our work.

## 2.2 Semantic Similarity

The problem of evaluating semantic similarity in NLP tasks has been studied in detail. Pedersen *et al.* (2004) present a survey of semantic similarity measures that rely on a *is-a* hierarchy. Zaki (2003) gives an extensive overview of similarity measures that use statistical techniques. For our research we have decided to use a statistical semantic similarity measure called PMI (Turney, 2001) using a terabyte corpus of unlabelled data. Turney explains that this similarity measure uses a logarithmic scale; therefore a value of zero indicates that two words are statistically independent, a high positive value indicates that two words tend to co-occur, and a negative value indicates that two words do not share the same lexical contexts. It has been shown that this measure of co-occurrence is a good estimator of semantic similarity (Turney, 2001; Terra and Clarke, 2003).

## 2.3 Lexical Cohesion

Lexical cohesion as presented by Halliday & Hasan (1976) is the notion that sentences and phrases of any sensible text will tend to be about the same thing. Indicators of cohesion are back-reference, conjunction and semantic word relations. Morris & Hirst (1991) have shown that it is possible to measure lexical cohesion automatically by identifying chains of related words that contribute to the continuity of lexical meaning. Systems that build *lexical chains* automatically have been implemented using *thesaural relations*. Lexical chain building systems have been realized using Roget's Thesaurus (Jarmasz & Szpakowicz, 2003) and WordNet (Silber & McCoy, 2000). Turney (2003) has shown that these thesaural relations can be captured effectively using statistical semantic similarity measures.

## 2.4 Keyphrase Extraction Evaluation

The literature presents two techniques for evaluating automatically generated keyphrases. The first is to use the author's keyphrases as a gold standard and calculate the overlap between the extracted keyphrases and the author's (Turney, 1999; Witten *et al.*, 1999). The main problem with this methodology is that the author's keyphrases are not always taken from the text. It is possible to make sure that the extracted keyphrases are correctly constructed noun phrases, named entities or are related to specific terminology by incorporating linguistic information and wordlists to the process, but it is extremely difficult to evaluate if a set of extracted keyphrases are adequate for a user to understand the gist of the text. Barker and Cornacchia (2000) propose another evaluation method which involves human judges. The judges must evaluate the quality of individual keyphrases and the entire set of keyphrases. Their research shows that using human judges for evaluation should be avoided as it is a difficult, time and energy-consuming process which does not always yield conclusive results (*ibid.*). We propose an alternative evaluation technique which relies on semantic similarity for comparing the author's keyphrases and the extracted keyphrases. We go beyond simple string matching without involving human judges.

## 3 Extraction and Informativeness

The experiments in this report are performed using the AFNJ document collection compiled by Turney (1999). It consists of 341 documents obtained by selecting 90 web pages from the Aliweb search engine, 35 web pages from the US Government's Federal Information Processing Standards, 141 web pages from NASA's Langley Research Center and 75 articles from five different academic journals. This corpus is of interest as every document in this collection is accompanied by a set of keyphrases generated by hand.

The first 50 documents of the AFNJ collection are used for our preliminary experiments. The smallest document contains 73 words, the largest 23,234. The average for the entire set is 1,159 words. We work on a small number of documents to be able to perform both a quantitative and a qualitative analysis based on thorough manual inspection of the results. This manual inspection is very important to correctly identify future research directions. The goal of this evaluation is two-fold: establish how keyphrases can be better presented to human readers, and identify outliers within the list of keyphrases.

We use Extractor to automatically generate a list of keyphrases from each of the first 50 documents in the AFNJ collection. This software can extract between 3 and 30 keyphrases from a text; we have decided to extract 15, as we want to identify a maximum of salient information while minimizing the amount of spurious information. Extractor presents the list of keyphrases according to its confidence in them; therefore the $2^{nd}$ should be more pertinent than the $10^{th}$. Manual inspection shows that this is not obviously apparent. In this research we consider all 15 to be of the same quality when refining the entire set.

Turney (2003) has performed experiments on specifying features which rely on semantic similarity to extract more cohesive sets of keyphrases using Kea. The work we present is different in that we refine the list once it has been generated rather than modifying the feature vectors of the extraction algorithm. Turney supposes that Kea's first four extracted keyphrases are of the best quality and attempts to identify keyphrases that are most similar to them, an assumption we do not make here.

### 3.1 Keyphrase Progression

Our first improvement is to determine the informativeness of each keyphrases. We calculate informativeness by counting in the Waterloo MultiText System (Terra and Clarke, 2003) the number of document in which a word or phrase occurs. We present 4 sets of keyphrases, extracted form documents 6, 8, 16 and 40 of the AFNJ collection, sorted from most general to most specific. The hit counts in the corpus are indicated in parentheses:

- *database* (2,854,665), *Lab* (1,304,478), *biology* (986,657), *Announcements* (833,477), *sequence* (718,236), *internet resources* (387,532), *Catalogs* (383,665), *molecular biology* (162,413), *Genome* (160,295), *ExPASy* (9,230), *biological software* (430), *enzyme databases* (130), *pioneer molecular biology* (0), *premier SwissProt* (0), *ExPASy Needs* (0).
- *fish* (1,127,309), *Oregon* (1,095,122), *biology* (986,657), *Molecular* (575,797), *Genetics* (411,768), *aquarium* (115,558), *model system* (13,080), *zebrafish* (6,648), *Molecular Data* (3,402), *Brachydanio rerio* (363), *Zebrafish Book* (184), *vertebrate developmental biology* (55), *Gilbert Lab Home* (2), *zebrafish servers* (2), *nosibork* (0).
- *food* (3,939,149), *parties* (1,531,690), *Christmas* (917,133), *catering* (201,722), *cigar* (77,191), *malt* (38,984), *Los Gatos* (24,293), *UPCOMING EVENTS* (16,878), *Libation* (5,074), *single malt* (4,891), *malt whisky* (2,559), *single malt whiskies* (254), *Pig Rig* (11), *Macallan Boycott* (0), *SCOTTISH NOTES* (0).
- *html* (7,199,750), *converter* (322,661), *MIT* (309,348), *scratch* (212,991), *hypertext* (200,241), *translator* (142,896), *latex* (116,169), *emacs* (64,104), *html formatting* (5,501), *html mode* (1,660), *writing hypertext* (604), *html helper mode* (596), *html modes* (112), *html formatting commands* (103), *exising latex code* (0).

These lists show that the extracted keyphrases can be very general, for example *food*, or very specific, for example *Gilbert Lab Home*.

### 3.2 Evaluation

We sort the keyphrases according to their informativeness to help the user peruse the list in a more natural manner and to identify outliers which should be removed. We have not evaluated if the re-ordering is beneficial to humans, but the sorting has yielded interesting results about the nature of extracted keyphrases.

The assumption that a keyphrase with a low hit count is informative reveals that there are incoherent keyphrases in the set. For example, the keyphrase *exising latex code* has a hit count of 0, suggesting that it is extremely informative, when in fact it is a spelling mistake. The keyphrase *ExPASy Needs*, again with a hit count of 0, is part of the phrase *ExPASy Needs Your Help!* Such keyphrases can be removed from the set to present a more coherent list to the reader.

Extractor attempts to select noun phrases, but sometimes includes verbs in the keyphrases. The keyphrases are between one and three words long, and the system does not rely on any linguistic information. Therefore it is common to see verb + noun or noun + verb combinations as keyphrases. A shallow parser or a tagger could be used to improve the results, but interestingly, a low hit count is an indicator of an incorrectly formed keyphrase, as it is not a common collocation. Table 1 shows some examples of errors found in the keyphrases extracted in our experiments as well as their hit counts.

| Document | Keyphrase | Type of error | Context | Hit count | Informativeness order | Extractor order |
|---|---|---|---|---|---|---|
| doc001 | Brews Success | V+N | … we're a restaurant that **Brews Success**. | 7 | 15 | 9 |
| doc002 | register web sites | V+N | If your job is to **register web sites** … | 88 | 14 | 7 |
| doc003 | Maintaing | Typo. | … got the resources to **maintaing** these pages | 1667 | 14 | 5 |
| doc005 | time spent referencing | N+V | Apollo demands results from **time spent referencing** | 0 | 15 | 8 |
| doc006 | ExPASy Needs | N+V | **ExPASy Needs** Your Help! | 0 | 15 | 15 |
| doc007 | porting utilities | V+ N | … promote the use of OS/2 by **porting utilities**… | 8 | 14 | 10 |
| doc010 | collection excludes letters | N+V+N | … this **collection excludes letters** transmitting … | 0 | 15 | 12 |
| doc011 | reporting progress | V+N | ... experimental bulletin board for **reporting progress** within … | 1390 | 11 | 5 |

**Table 1:** Examples of keyphrase extraction errors and their estimated informativeness

Sorting the list of extracted keyphrases according to their informativeness and removing those that fall below a certain threshold seems to be a better technique rather than simply relying on Extractors rankings. This is presented by the discrepancies in orderings in Table 1.

The author-selected keyphrases for the sample of the AFNJ collection that we study is an indicator of what ideal lists should look like. For these 50 documents, the authors have selected as little as 2, and as much as 15 keyphrases, the average being 6.63. In 28.3% of the cases, the keyphrase is not in the text, and in another 4.3% it is present but in a different form. Using this sample, in the best of worlds we can never obtain more than a 72% overlap between the author's keyphrases and the automatically extracted ones. In state-of-the-art systems, sets of 15 keyphrases obtain a maximum overlap of 2.5 keywords per set (Witten *et al.*, 1999; Turney, 2003), although this does not indicate that the computer selected keyphrases are inadequate. We therefore propose to use semantic similarity to calculate the overlap between sets instead of string matches.

We use the PMI measure (Turney, 2001) to estimate semantic similarity. It uses a logarithmic scale, and is based on co-occurrence. The probabilities for the measure are calculated using the Waterloo MultitText System using the following formula:

$$PMI(w_1, w_2) = \log_2 \left( \frac{p(w_1 \cap w_2)}{p(w_1)p(w_2)} \right)$$

**Figure 1:** PMI formula (Turney, 2001)

There are several ways of estimating PMI scores but Terra and Clarke (2003) have shown their corpus to be superior for this task.

We can now identify which Extractor keyphrases are the most similar to the author's by calculating the average similarity between sets. We show in bold the author's keyphrases from documents 6, 8, 16 and 40 in their original order and specify between brackets the extracted keyphrases which are the most similar to the author's (highest positive PMI score). The extracted keyphrases are sorted from general to specific:

- **Biochemistry**, **Molecular** (*molecular biology*, *ExPASy*, *biological software*), **Cellular**, **Developmental**, **Organismal** (*Lab*), **Evolutionary**, **Biology** (*biology*, *molecular biology*), **Harvard**, **Genomics** (*sequence*, *Genome*, *enzyme databases*)

| Document | Set similarity (*all 15 keyphrases*) | Set similarity (*keyphrases < 100 hit counts removed*) | Set similarity (*5 least frequent keyphrases removed*) | Set similarity (*3 least frequent and 2 most frequent keyphrases removed*) |
|---|---|---|---|---|
| doc006 | -2.99 | -0.66 | -0.26 | -0.59 |
| doc008 | -0.84 | 1.19 | 1.10 | 2.00 |
| doc016 | -3.50 | -0.89 | -1.34 | -0.80 |
| doc040 | -1.08 | -0.20 | -0.15 | 0.45 |
| Average over 50 documents | -2.73 | -1.95 | -1.78 | -1.86 |

**Table 2:** Impact of list reduction heuristics on average semantic similarity

- **Genetics** (*Genetics*), **Biology**, **Neuroscience** (*Zebrafish Book*), **Developmental Biology** (*biology, Molecular, model system, zebrafish, Molecular Data, Brachydanio rerio, vertebrate developmental biology, zebrafish servers*)
- **Restaurant** (*food, catering*), **Irish**, **Ireland**, **Sports**, **Bar**, **Pub**, **Whisky** (*malt whisky*), **Malt** (*cigar, malt, Libation, single malt , single malt whiskies*)
- **HTML** (*html, translator, latex, html formatting, html mode, html helper mode, html formatting commands*), **hypertext** (*MIT, hypertext, emacs, writing hypertext*) , **writing**, **authoring** (*html modes*)

The overlap that is found between the author's and the automatically generated keyphrases is higher compared to string matching. We notice that certain author's keyphrases are dominant; they summarize better several extracted keyphrases, for example *Developmental Biology* in document 8.

The semantic similarity using the PMI measure returns values which cannot be used in absolute terms and therefore cannot perform a system`s evaluation in isolation. It is nevertheless quite valuable as it allows for a repeatable evaluation method between keyphrase extraction systems or between versions of the same system; the more similar the extracted set of keyphrases to the author's, the better the system.

We have tested three heuristics to reduce the list of keyphrases and increase the overall semantic similarity. The first is to remove all keyphrases with a hit count less than 100; the second is to remove the 5 least frequent keyphrases; the third is to remove the 3 least frequent and the 2 most frequent, thus the most specific and the most general keyphrases. The results are presented in Table 2.

The heuristics used to reduce the size of the keyphrase list increase the semantic similarity. Although the PMI score is still negative, this is a satisfactory result as PMI is used as a relative value of semantic similarity and not an absolute one.

We can now study Extractor's ordering of keyphrases according to the average PMI scores. Taking the set of the first 5, and 10 extracted keyphrases the average score is of -1.19 and -2.20 respectively. Taking the first 5 keyphrases leads to a good similarity with the author`s set. On the other hand taking the set of the first 10 keyphrases is worse than the set obtained by removing 5 keyphrases using our list reduction heuristics.

## 4 Keyphrase Clustering

We have discussed presenting keyphrases sorted from general to specific. This is interesting if a document is only about a single topic. We know that this is not a realistic assumption. Therefore we propose to identify the various topics in a text by clustering the keyphrases according to their semantic similarity.

### 4.1 Keyphrase Cluster Generation

We use the same 50 documents and Extractor for the keyphrase clustering experiment. The keyphrases are clustered using a standard bottom-up clustering algorithm. A feature vector for every keyphrase is defined by taking the semantic similarity values between itself and the fourteen remaining keyphrases. We calculate the cosine of the angle between all pairs of vectors to determine the similarity of a keyphrase to another relative to the entire set of extracted keyphrases from the document. We set the initial number of clusters to five, which is about the limit of the human capacity for processing the information (Miller, 1959). These clusters allow viewing keyphrases by topic.

| Text | Clusters |
|---|---|
| doc008 | (1) Molecular Data |
| | (2) zebrafish servers |
| | (3) Brachydanio rerio, vertebrate developmental biology |
| | (4) fish, Oregon, biology, Molecular, Genetics, aquarium, model system, zebrafish, Zebrafish Book |
| | (5) Gilbert Lab Home, nosibork |
| doc021 | (1) Business, Education, Entertainment, Computers |
| | (2) Ireland, Swift, seen Swift |
| | (3) Magazines |
| | (4) Art, Film, Literature, Archaeology |
| | (5) Government, Genealogy, Festivals |

**Table 3:** Two examples of keyphrase clusters

| | Heuristic A: 3 smallest clusters removed | Heuristic A applied only when the 2nd cluster >= 3 keyphrases | Heuristic A applied only when the 2nd cluster >= 4 keyphrases |
|---|---|---|---|
| **# of documents** | 50.00 | 33.00 | 15.00 |
| **Similarity before removing keyphrases** | -2.73 | -2.61 | -2.29 |
| **Similarity keeping largest cluster** | -1.72 | -1.63 | -1.84 |
| **Similarity after removing keyphrases** | -2.32 | -2.44 | -2.12 |
| **Average size of largest cluster** | 8.10 | 7.50 | 6.60 |
| **Average number of keyphrases** | 11.10 | 11.10 | 10.80 |

**Table 4:** Impact of removal of clusters on average semantic similarity

### 4.2 Cluster Evaluation

Table 3 presents two sets of five clusters for documents 8 and 21 of the AFNJ collection. The lists of keyphrases within the clusters are further sorted according to their information content. The cluster *fish*, *Oregon*, *biology*, *Molecular*, *Genetics*, *aquarium*, *model system*, *zebrafish*, *Zebrafish Book* contains most of the keyphrases extracted from document 8. This suggests that the extracted keyphrases are very cohesive. In contrast, the clusters of document 21 are well balanced. Document 8 is more representative of this small sample of the AFNJ collection as it is shown in Table 4. In the 50 documents, there are on average 8.1 keyphrases in the largest cluster, which represents 54% keyphrases in 20% of clusters; 11.1, or 73%, are found in the first two largest clusters. In 66% of documents, the second largest cluster contains 3 or more keyphrases, and in 30% of documents, the second largest cluster contains 4 or more keyphrases. Further experimentation is required to confirm if Extractor keyphrases are generally cohesive or if the sample of texts biases the results.

We have not performed a qualitative analysis of these clusters using human judges, and therefore do not know how this representation affects the reader, but we have performed a quantitative evaluation using the PMI similarity measure. Our results seem to indicate that the extracted keyphrases are cohesive, as they cluster mostly in one or two large sets. This is confirmed by the increase of the average PMI score with author's keyphrases when only the largest set is kept. This is the most effective keyphrase reduction technique, even when compared with the heuristics which rely on informativeness. The other keyphrase reduction techniques which rely on clustering are not as effective as the ones which rely on informativeness.

## 5 Conclusion

We looked at two methods for improving the organization of lists of keyphrases and removing outliers. Informativeness estimated using hit counts on a terabyte-sized corpus is a simple technique which performs well for identifying outliers. Very infrequent keyphrases, those with hit counts less than 100, often are mostly erroneous and do not repre-sent specific terms as we initially proposed. A sophisticated technique relying on the PMI similarity measure to cluster the keyphrases according to their topics allows, for the short texts in our collection, to identify one large cohesive cluster that is more similar to the author's than the overall set. Other smaller clusters can be seen as less significant and even outliers.

Both information content and clustering seem promising avenues of research which we plan to pursue with the goal of improving the organization and the quality of lists of keyphrases.

Our major contribution is to propose a novel way for evaluating sets of keyphrases using semantic similarity. Traditional techniques of comparing the overlap between a set of automatically generated keyphrases and a gold standard using string matching or relying on human judges are problematic. By measuring the average semantic similarity between the set of extracted keyphrases and the gold standard it is possible to give a relative score which can be used to compare various keyphrase extraction algorithms.

## Acknowledgements

We are most grateful to Egidio Terra, Charlie Clarke, and the School of Computer Science of the University of Waterloo, for giving us a copy of the Waterloo MultiText System. We would also like to thank Peter Turney for his comments and help at various stages of this research.

## References

Barker, K. and N. Cornacchia. 2000. Using Noun Phrase Heads to Extract Document Keyphrases. *Proceedings of the Thirteenth Canadian Conference on Artificial Intelligence* (*AI 2000*). Montréal, Canada, 40 – 52.

Clarke, C.L.A., G.V. Cormack and F.J Burkowski. 1995. An algebra for structured text search and a framework for its implementation. *The Computer Journal,* 38(1): 43 – 56.

Clarke, C.L.A. and G.V. Cormack. 2000. Shortest substring retrieval and ranking. *ACM Transactions on Information Systems (TOIS)*, 18(1): 44 – 78.

Copek, T., N. Japkowicz and S. Szpakowicz. 2002. Text Summarization as Controlled Search. *In Proceedings of the 15th Canadian Conference on Artificial Intelligence (AI 2002)*, Calgary, Canada, May, 268 – 280.

Frank, E., G.W. Paynter, I.H. Witten, C. Gutwin and C.G. Nevill-Manning. 1999. Domain-specific keyphrase extraction. *Proceedings of the Sixteenth International Joint Conference on Artificial Intelligence (IJCAI-99),* California, 668-673.

Halliday, M. and R. Hasan. 1976. *Cohesion in English*. Longman.

Jarmasz, M. and S. Szpakowicz. 2003. Not As Easy As It Seems: Automating the Construction of Lexical Chains Using Roget's Thesaurus. *In Proceedings of the 16th Canadian Conference on Artificial Intelligence (AI 2003)*, Halifax, Canada, June, 544–549.

Miller, G.A. 1956. The Magical Number Seven, Plus or Minus Two: Some Limits on Our Capacity for Processing Information. *In The Psychological Review*, 63: 81 – 97.

Morris, J. and G. Hirst. 1991. Lexical Cohesion Computed by Thesaural Relations as an Indicator of the Structure of Text. *Computational linguistics*, 17:1, 21 – 48.

Pedersen, T. S. Patwardhan, and J. Michelizzi. 2004. WordNet::Similarity - Measuring the Relatedness of Concepts. *In Proceedings of Fifth Annual Meeting of the North American Chapter of the Association for Computational Linguistics (NAACL-04)*, Boston, MA, May, 267-270.

Shannon, C. 1948. A Mathematical Theory of Communication. *The Bell System Technical Journal*, 27: 379 – 423, 623 – 656.

Silber, H.G., K.F. McCoy. 2000. Efficient text summarization using lexical chains. *Intelligent User Interfaces*, 252 – 255.

Terra, E. and C.L.A. Clarke. 2003. Frequency estimates for statistical word similarity measures. *In Proceedings of the Human Language Technology and North American Chapter of Association of Computational Linguistics Conference 2003 (HLT/NAACL 2003)*, Edmonton, Canada, 244 – 251.

Turney, Peter D. 1999. Learning to Extract Keyphrases from Text. National Research Council, Institute for Information Technology, Technical Report ERB-1057.

Turney, P.D. 2001. Mining the Web for Synonyms: PMI-IR vs. LSA on TOEFL. *Proceedings of the 12th European Conference on Machine Learning (ECML-2001)*, Freiburg, Germany, 491 – 502.

Turney, P.D. 2003. Coherent Keyphrase Extraction via Web Mining. In *Proceedings of the Eighteenth International Joint Conference on Artificial Intelligence (IJCAI-03),* Acapulco, Mexico, 434 – 439.

Witten, I.H., G.W. Paynter, E. Frank, C. Gutwin and C.G. Nevill-Manning. 1999. KEA: Practical Automatic Keyphrase Extraction. *In Proceedings of the Fourth ACM Conference on Digital Libraries (DL'99)*, Berkeley, USA, August, 254 – 256.

Zaki, S. 2003. Exploring Word and Sentence Similarity in Corpus. M.A.Sc. Thesis. School of Information Technology and Engineering, University of Ottawa.